\documentclass[runningheads]{llncs}
\usepackage[T1]{fontenc}
\usepackage{graphicx}
\usepackage[numbers,sort&compress]{natbib}
\usepackage{float}
\usepackage[caption=false]{subfig}
\usepackage{amsmath} 
\begin{document}
\raggedbottom 
\title{Trusting Right Predictions for Wrong Reasons: A LIME Based Analysis of Deep Learning Interpretability in Lung Cancer Diagnosis}
\titlerunning{False Alarm Analysis Using XAI} 
%
%
\author{Samarpan Poudel \and Vladislav D Veksler}
\authorrunning{S. Poudel \& V. D. Veksler} 
\institute{Caldwell University \\ School of Business and Computer Science \\
Caldwell, NJ 07006 \\
\email{\{spoudel3,vveksler\}@caldwell.edu} }
\maketitle {}             
\begin{abstract}

Lung cancer is the leading cause of cancer related mortality, with current statistics indicating approximately 2.5 million new cases and 1.8 million deaths annually, making accurate and trustworthy diagnosis a clinical priority. Although deep learning models have demonstrated strong classification performance on this task, the field has largely evaluated them on accuracy alone, leaving the question of how and where these models make their decisions largely unexamined. This study compares three architecturally distinct models, a Convolutional Neural Network, a pretrained ResNet50, and a Vision Transformer built from scratch, trained on the IQ-OTH/NCCD lung cancer CT dataset, and applies Local Interpretable Model-Agnostic Explanations (LIME) to analyze their decision making patterns. Beyond standard performance evaluation, this work introduces a dual correlation framework that computes both prediction agreement and explanation agreement across all model pairs. The results show that all three models achieved strong classification performance, with ResNet50 reaching 98.61\% accuracy, the CNN 97.91\%, and the Vision Transformer 93.75\%, with ROC-AUC scores of 0.99 across all three. Pairwise prediction correlations exceeded 0.99 across all model pairs, yet LIME explanation correlations remained below 0.26, revealing that near identical predictions are being produced through fundamentally different visual reasoning processes. Furthermore, LIME analysis identified a consistent spatial pattern in misclassified samples. Incorrect predictions were systematically associated with model attention directed outside the lung parenchyma, while correct predictions concentrated within it. These findings demonstrate that prediction agreement is a poor proxy for reasoning consistency, and that interpretability evaluation must be treated as an independent validation criterion alongside predictive performance in clinical AI systems.

\keywords{ Explainable AI, False Alarms, Correlation Analysis}

\end{abstract}

\section{INTRODUCTION}

Lung Cancer remains one of the leading causes of cancer related deaths worldwide with approximately 1.8 million deaths in 2020 alone representing nearly 18\% of all cancer deaths globally \citep{sung2021global}. Despite advances in treatments, the five year survival rate has dropped below 20\% in most developed countries, a number that improves significantly when the disease is caught at an early stage \citep{hattori2017importance, fu2019distinct}. Computed tomography (CT) imaging scans have become the primary modality for early lung cancer screening. Over the past decade, deep learning has emerged as a compelling tool for automating the interpretation of those scans. Convolutional neural networks have demonstrated strong performance in this domain, with multiple studies reporting classification accuracies exceeding 95\% on benchmark datasets \citep{pandian2022detection, shariff2025optimizing}. Recently, Vision Transformers, first introduced by Dosovitskiy et al. \citep{dosovitskiy2020image} as a patch-based attention mechanism adapted from natural language processing, have entered the medical imaging domain and shown competitive results, sparking growing interest in comparing these architectures against established convolutional designs \citep{faizi2025deep, gai2024comparing}.

However, comparative studies rarely focus on metrics beyond accuracy and signal detection. A model that achieves 98\% accuracy on a test dataset is typically considered validated, and if two models agree on their predictions, that agreement is treated as a further signal of reliability. What these evaluations miss is the question of whether the model is making its decisions for the right reasons. A model can score 98\% accuracy while making its decisions based on image features that have nothing to do with pulmonary pathology. Deep learning models are inherently opaque, and high accuracy does not guarantee that a model has learned clinically relevant features. Zech et al. \citep{zech2018variable} reported a directly comparable pattern in chest radiograph classifiers, where models attending to image periphery rather than pulmonary tissues achieved strong benchmark performance while failing to generalize across institutions. Despite this finding, most studies still only focus on performance metrics
(with some notable exceptions; e.g., \citep{panboonyuen2026seeingisntbelievinganalysis, dwivedi2023explainable}).
Furthermore, very few studies have systematically compared whether architecturally distinct models that agree on predictions also agree on the image regions they use to reach those predictions \citep{raghu2021vision}. This work addresses these gaps by systematically quantifying spatial explanation agreement across diverse architectures.

Using a custom CNN, a pretrained ResNet50, and a Vision Transformer trained on the IQ-OTHNCCD lung cancer CT dataset, this study applies LIME explainability analysis to examine the spatial attention patterns associated with correct versus incorrect predictions, and computes both prediction correlation and explanation correlation across all three model pairs. The study is purposely limited to a single dataset and explainable technique LIME for easy and correct architecture comparisons. The findings have practical implications for how clinical AI is validated, and for what it actually means when multiple models agree on their predictions.

\section{Literature Review}
Convolutional neural networks have dominated lung cancer classification from CT imaging for years, with ResNet based architectures consistently reporting strong performance on binary malignancy detection tasks. Rajasekar et al. \citep{rajasekar2023lung} evaluated six deep learning architectures including CNN, ResNet50, VGG16, and InceptionV3 on CT and histopathological images, finding CNN based models achieving accuracy above 97\%. Transfer learning through pretrained models such as ResNet has shown particular promise under limited data conditions, with fine tuned convolutional architectures consistently outperforming models trained from scratch by meaningful margins \citep{hossain2022transfer}. Kumar et al. \citep{kumar2024vision} applied a pretrained Vision Transformer as a feature extractor on the LC25000 histopathology dataset, freezing ViT layers and adding a classification head, achieving 98.84\% accuracy at a patch size of $16 \times 16$ demonstrating ViTs can effectively capture complex spatial relationships in lung cancer tissue images.

The opacity of these models has pushed explainability methods into the conversation, with LIME, GradCAM, and SHAP becoming standard additions to medical imaging papers. In cancer work specifically, LIME has been applied to highlight regions in CT images that contribute most to model predictions, with the stated aim of boosting clinician trust and supporting diagnostic decision making \citep{ukwuoma2025enhancing}. Wani et al. \citep{wani2024deepxplainer} paired a hybrid CNN-XGBoost model with SHAP based explanation and showed that high accuracy models can be made more transparent without sacrificing performance. Tomassini et al. \citep{tomassini2023cloud} went further by providing dynamic heatmap visualizations for individual CT scans highlighting most influential lung regions in the model decision making process. These are useful contributions, but they all share a common limitation of explainability that is applied to one model at a time rather than used to compare how different models reason about the same images.

Existing research on AI model failures makes this specific gap even more significant. Deep learning models in medical imaging can rely on shortcut learning by exploiting spurious correlations not causally related to the target task, which poses real risks in clinical settings where models must generalize across institutions. Geirhos et al. \citep{geirhos2018imagenet, geirhos2020shortcut} demonstrated that convolutional neural networks trained on image data primarily focus on surface level statistical patterns over structurally meaningful features, a tendency that appears in CT imaging as vulnerability to scanner, image boundaries, and non-pulmonary tissue. No study to date has quantified whether models agreeing on predictions also agree on explanations, or whether misclassification in lung CT analysis is spatially tied to outside lung attention. This paper addresses both gaps.

\section{Methodology}
\subsection{Dataset Preparation}
The dataset used in this study was obtained from the Iraq-Oncology Teaching Hospital/National Center for Cancer Diseases (IQ-OTH/NCCD) lung cancer dataset, which is publicly available on Mendeley Data. The original dataset consists of 1,190 CT scans which were split into training, validation and testing using an 80:10:10 ratio. Data augmentation was applied to the training dataset in order to improve model generalization, while the validation and testing sets were kept unchanged to ensure unbiased evaluation. After augmentation, the training dataset contained 2,176 images. Specifically, we leveraged the \textit{Augmentor} library to execute a co-augmentation pipeline that applied randomized rotations $\pm 5^\circ$, horizontal and vertical flips, and spatial zooming directly to the raw images \citep{bloice2017augmentor}.\par
\begin{figure}[htbp]
    \centering
    \begin{minipage}{0.33\textwidth}
        \centering
        
        \includegraphics[width=\textwidth]{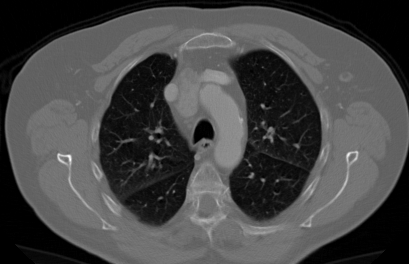}
        {\small (a) Positive}\par\medskip
    \end{minipage}
    \hspace{0.05\textwidth} 
    \begin{minipage}{0.3\textwidth}
        \centering
        
        \includegraphics[width=\textwidth]{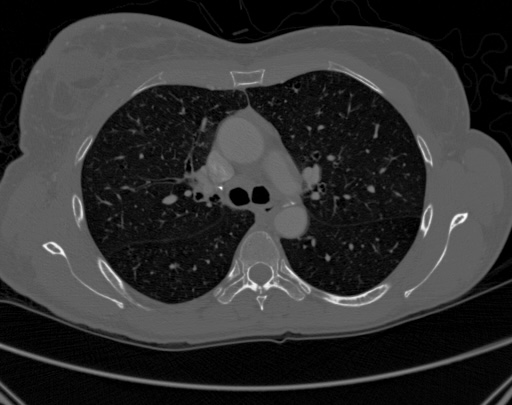}
        {\small (b) Negative}\par\medskip
    \end{minipage}
    \vspace{2mm}
    \caption{Cancer and Non Cancer Images from Dataset}
    \label{fig:sample_images}
\end{figure}

The dataset was loaded into a Jupyter Notebook environment using TensorFlow’s Keras utility image\_dataset\_from\_directory. All images were resized to $256 \times 256$ pixels, with a batch size of 32. Sample images from the dataset are displayed in Fig. \ref{fig:sample_images}.

\subsection{Model Architecture and Training}
\subsubsection{CNN}
\begin{figure}[H]
    \centering
    \includegraphics[width=0.6\textwidth]{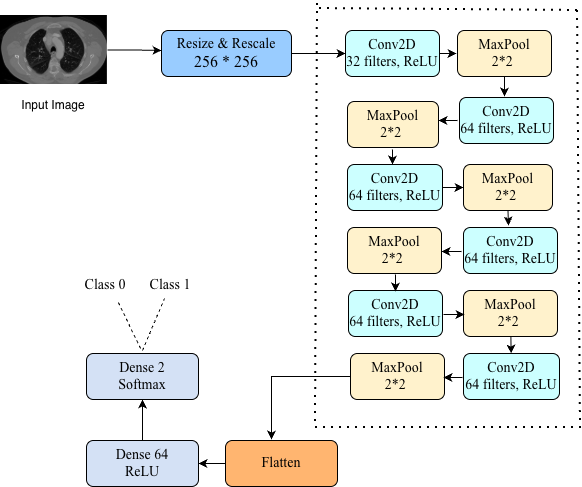}
    \caption{CNN Architecture}
    \label{fig:CNN_architecture}
\end{figure}

A Convolutional Neural Network (CNN) was designed and trained from scratch on the lung cancer dataset. The architecture consists of six convolutional layers with $3 \times 3$ kernels and ReLU activation, each followed by a max-pooling layer for spatial downsampling. The network begins with 32 filters in the first convolutional layer and increases to 64 filters in subsequent layers, enabling the model to learn progressively more complex features. After feature extraction, the output is flattened and passed through a fully connected dense layer with 64 neurons, followed by a softmax output layer for binary classification as shown in Fig. \ref{fig:CNN_architecture}.

\subsubsection{Pretrained ResNet50}
\begin{figure}[H]
    \centering
    \includegraphics[width=0.8\textwidth]{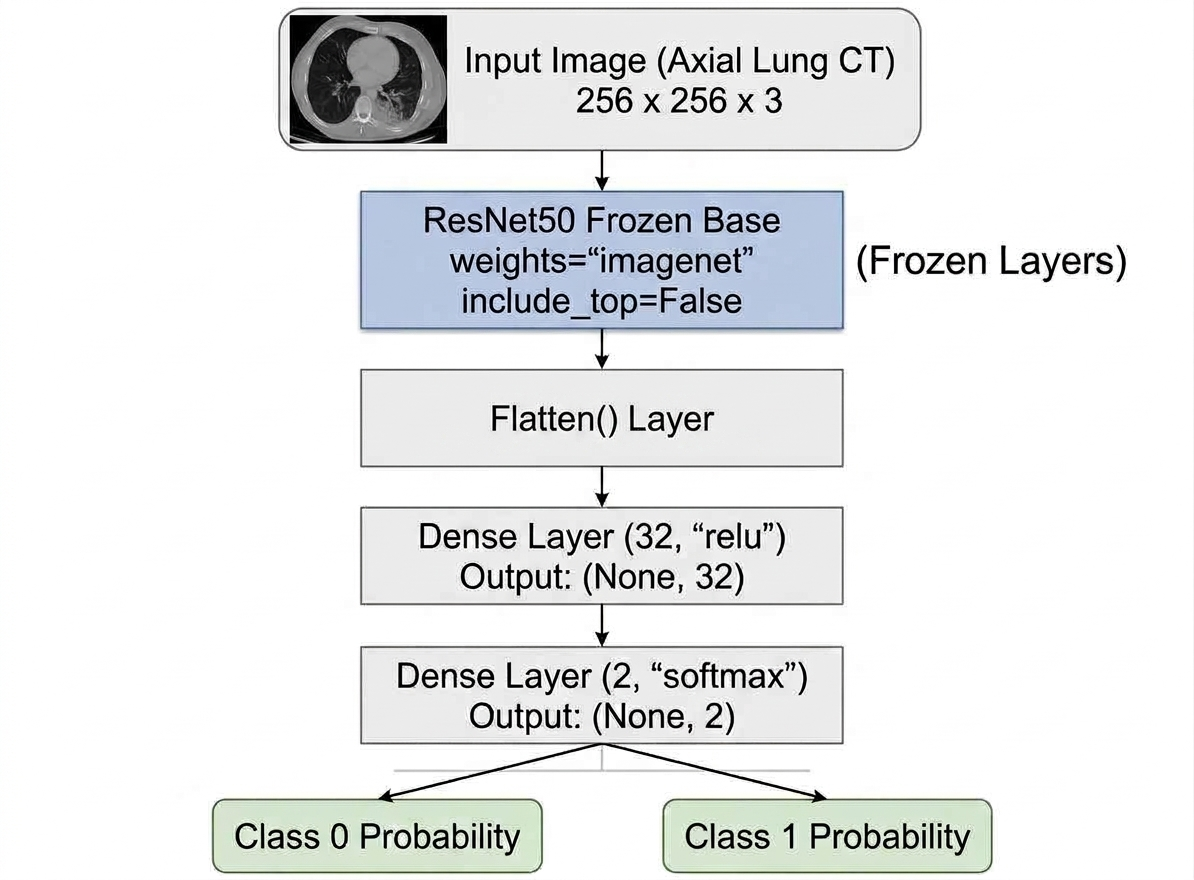}
    \caption{ResNet50 Architecture}
    \label{fig:ResNet50_architecture}
\end{figure}

We used a ResNet50 pretrained on ImageNet as a transfer learning baseline. All outside trainable parameters were frozen and the model was trained on the same dataset. At the end, the output was flattened and passed through a dense layer with 32 neurons and relu activation followed by two neuron dense layers with softmax activation as shown in Fig. \ref{fig:ResNet50_architecture}.

\subsubsection{Vision Transformer}
\begin{figure}[H]
    \centering
    \includegraphics[width=0.9\textwidth]{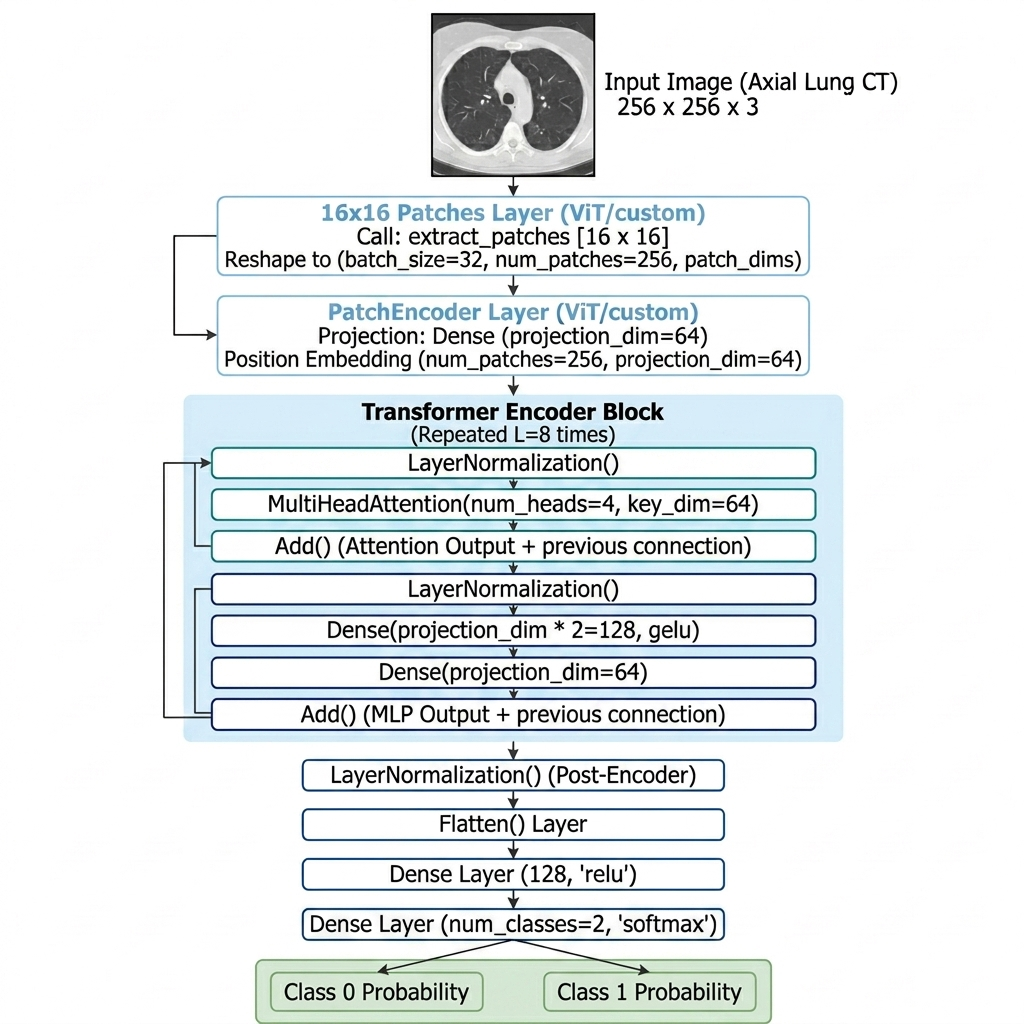}
    \caption{Transformer Architecture}
    \label{fig:ViT_architecture}
\end{figure}

A Vision Transformer (ViT) model was built to study how well transformer based methods work for lung cancer classification. Instead of using convolutions, the model splits each $256 \times 256$ image into $16 \times 16$ patches and treats them as a sequence so it can learn global relationships using attention.
Each patch is converted into a 64-dimensional embedding and positional information is added so the model knows where each patch comes from. The sequence is then passed through 8 transformer encoder layers with multi head attention, normalization, residual connections, and feed forward networks.
This setup helps the model focus on relationships across the entire image rather than local regions alone, which is useful for medical scans. The final output is flattened and passed through a dense layer with 128 neurons and then a softmax layer for binary classification. The model was trained using the same dataset and settings as the CNN and ResNet models for fair comparison. Figure \ref{fig:ViT_architecture} provides a visual overview of the Transformer architecture.

\subsubsection{Model Training and Performance Metrics}
All three models were trained in identical conditions. Adam optimizer and Sparse Categorical Cross Entropy was compiled into the models before training. The models were instructed to train to 100 epochs by adding early stopping which prevents overfitting. Model performances were determined using six evaluation metrics: accuracy, precision, recall, F1 score, ROC-AUC and Sparse categorical cross-entropy loss. They were computed as follows:

\begin{equation}
\text{Accuracy} = \frac{\text{TN} + \text{TP}}{\text{TP} + \text{TN} + \text{FP} + \text{FN}}
\end{equation}

\begin{equation}
\text{Precision} = \frac{\text{TP}}{\text{TP} + \text{FP}}
\end{equation}

\begin{equation}
\text{Recall} = \frac{\text{TP}}{\text{TP} + \text{FN}}
\end{equation}

\begin{equation}
\text{F1} = 2 \times \frac{\text{Precision} \times \text{Recall}}{\text{Precision} + \text{Recall}}
\end{equation}

\begin{equation}
\text{AUC} = \int_{0}^{1} \text{TPR}(\text{FPR}^{-1}(x)) \, dx
\end{equation}

\begin{equation}
\text{Loss} = - \frac{1}{N} \sum_{i=1}^{N} \log(\hat{y}_{i, c_i})
\end{equation}

\subsection{Explainable AI}
\begin{figure}[H]
    \centering
    \includegraphics[width=0.4\textwidth]{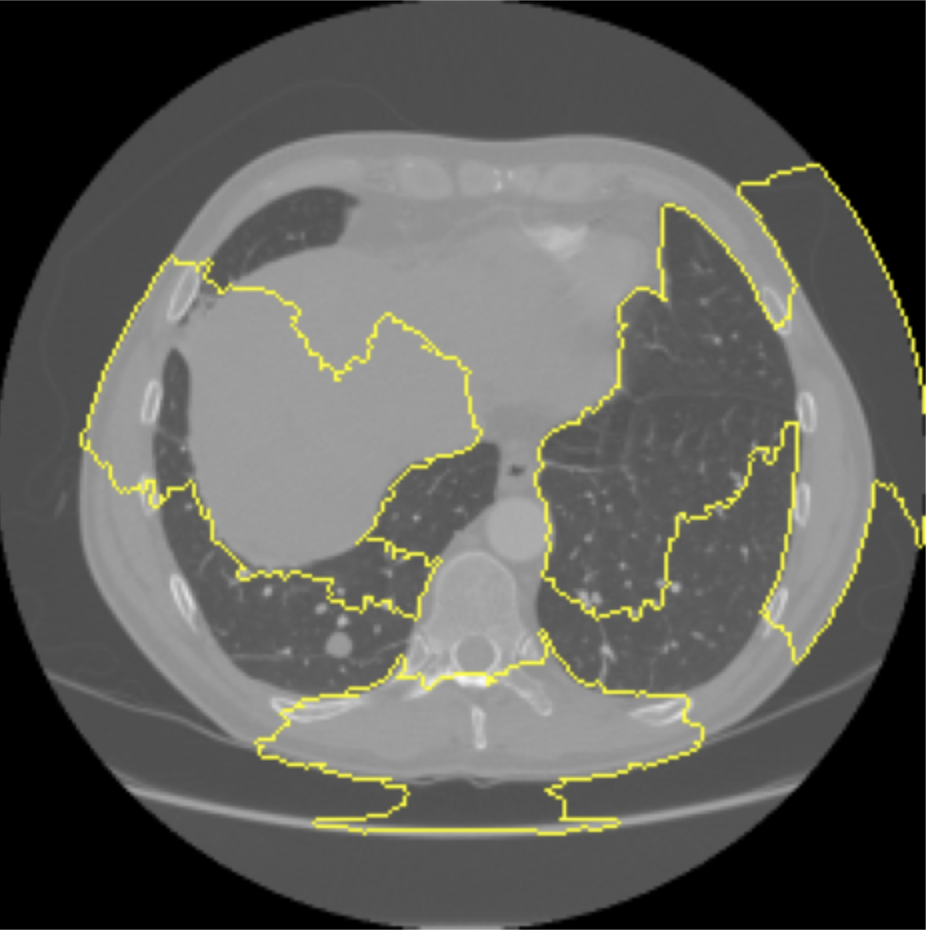}
    \caption{Sample Image Explanation}
    \label{fig:sample_image_explanation}
\end{figure}

The explainable AI (XAI) technique was applied to identify the models decision making approach. After training, predictions were generated on test set, and the correct and incorrect predictions of each model were stored in a directory for further testing. Local Interpretable Model-Agnostic Explanations (LIME) was chosen to perform the XAI experiments. To ensure reproducibility, the local surrogate was evaluated using 1,000 perturbed samples with default Quickshift superpixel segmentation.\ To ensure valid spatial comparison across models, the same fixed segmentation mask was applied to all three models during the LIME analysis. Explanation features of the models in a particular image were drawn using Lime Image Explainer and visualized using the matplotlib library. Figure \ref{fig:sample_image_explanation} illustrates a sample LIME explanation.

The local behavior of the black box architecture is mathematically approximated by minimizing the core LIME objective function:

\begin{equation}
\xi(x) = \arg\min_{g \in G} \mathcal{L}(f, g, \pi_x) + \Omega(g)
\end{equation}

Where the components and symbols are defined as follows:
\begin{itemize}
    \item $\xi(x)$: The final local explanation generated for the specific target image instance $x$.
    \item $f$: The complex, non-linear deep learning model being evaluated (e.g., Custom CNN, ResNet50, or ViT).
    \item $g$: The simple, inherently interpretable local surrogate model (a Ridge regression linear model) chosen from the class of explainable models $G$.
    \item $\mathcal{L}(f, g, \pi_x)$: The local loss function, which measures how closely the explanations of the surrogate model $g$ approximate the continuous predictions of the black-box model $f$.
    \item $\pi_x$: The proximity measure or similarity kernel that weights the generated perturbations based on their Euclidean distance to the original instance $x$, ensuring the surrogate model prioritizes local fidelity.
    \item $\Omega(g)$: The complexity penalty that acts as a regularization constraint, restricting the number of highlighted features (superpixels) to keep the final visual explanation human-interpretable.
\end{itemize}

\subsection{Correlation Analysis}

The output of the Lime Image Explainer was stored as a 2D numpy array using explanation segments as superpixel IDs. In a similar way, the output of the softmax layer was also stored as a prediction probability. Pearson correlation method was used to identify the correlation of the prediction and explanation similarity between all the models. The Pearson correlation is calculated as follows:
\begin{equation}
r = \frac{n\sum xy - (\sum x)(\sum y)}{\sqrt{\vphantom{1} n\sum x^2 - (\sum x)^2} \sqrt{\vphantom{1} n\sum y^2 - (\sum y)^2}}
\end{equation}

\section{Results}

\subsection{Classification Performance}

All three models achieved strong performance on IQ-OTH/NCCD lung cancer dataset. ResNet50 achieved highest accuracy at 98.61\% with 0.99 F1 score and ROC-AUC followed by CNN at 97.91\% with 0.98 F1 score and ROC-AUC of 0.99. The vision transformer achieved an accuracy of 93.75\% with F1 score of 0.94 and ROC-AUC of 0.99. While ResNet50 and Custom CNN outperformed Vision Transformer in terms of accuracy, ROC-AUC scores show that all three models are actually comparable when it comes to separating the data. 
Results are shown in Table \ref{tab:model_results}.

\begin{table}[htbp]
\centering
\caption{Classification Performance Metrics Across Architectures}
\label{tab:model_results}
\setlength{\tabcolsep}{4pt}       
\renewcommand{\arraystretch}{1.2} 
\begin{tabular}{lccccc}
\hline
\textbf{Models} & \textbf{Accuracy (\%)} & \textbf{F1 Score} & \textbf{Precision} & \textbf{Recall} & \textbf{ROC-AUC} \\ \hline
Custom CNN & 97.91                  & 0.98              & 0.98               & 0.98            & 0.99    \\ 
ViT        & 93.75                  & 0.94              & 0.94               & 0.94            & 0.99    \\ 
ResNet50   & 98.61                  & 0.99              & 0.99               & 0.99            & 0.99    \\ \hline
\end{tabular}
\end{table}

\subsection{Spatial Attention Patterns in Misclassified Samples}

Despite high classification accuracy, LIME explanation maps exposed a consistent pattern in misclassified samples across all three models. Predictions were most likely to be correct when the model concentrated most of its attention on superpixels inside the lung region. In contrast, when attention is shifted toward anatomical regions such as chest walls or image backgrounds, the probability of misclassification increases.

\begin{figure}[H]
    \centering
    \begin{minipage}{0.42\textwidth}
        \centering
        
        \includegraphics[width=\textwidth]{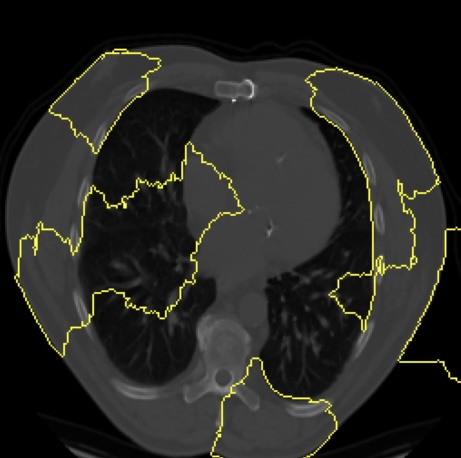}
        {\small (a) Correct Classification}\par\medskip
        \label{fig:CNN_Correct_Case}
    \end{minipage}
    \hspace{0.05\textwidth} 
    \begin{minipage}{0.42\textwidth}
        \centering
        
        \includegraphics[width=\textwidth]{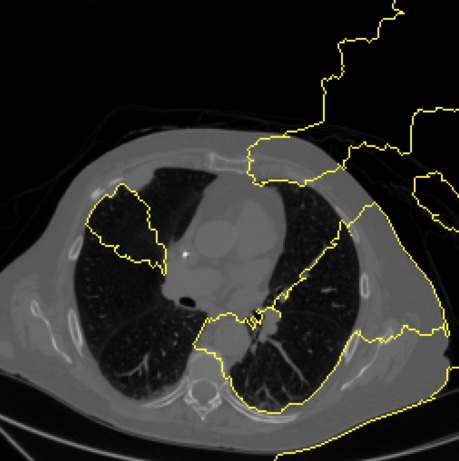}
        {\small (b) Incorrect Classification}\par\medskip
        \label{fig:CNN_Incorrect_Case}
    \end{minipage}
    \vspace{2mm}
    \caption{Correct and Incorrect Predictions by Custom CNN.}
    \label{fig:CNN_Case}
\end{figure}

In all false cases, the highlighted superpixels were predominantly located outside the lung region, suggesting that the model responded to image features rather than pathologically meaningful tissues. This pattern was observed consistently across all three architectures, showing misclassification is not random but is systematically linked to where in the image the model focuses during decision making. Models biased toward non-pulmonary regions carry a meaningful clinical risk, as they may produce false outputs when used on different patient groups.

\begin{figure}[H]
    \centering
    \begin{minipage}{0.42\textwidth}
        \centering
        
        \includegraphics[width=\textwidth]{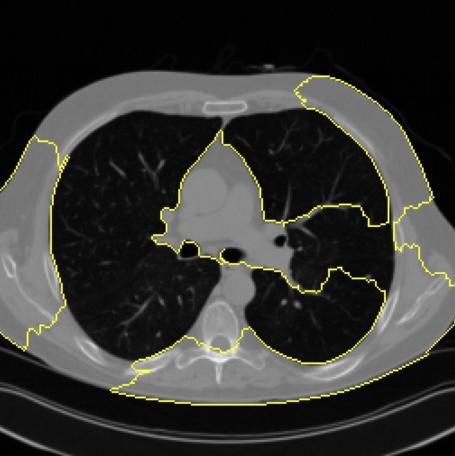}
        {\small (a) Correct Classification}\par\medskip
        \label{fig:Resnet50_Correct_Case}
    \end{minipage}
    \hspace{0.05\textwidth} 
    \begin{minipage}{0.39\textwidth}
        \centering
        
        \includegraphics[width=\textwidth]{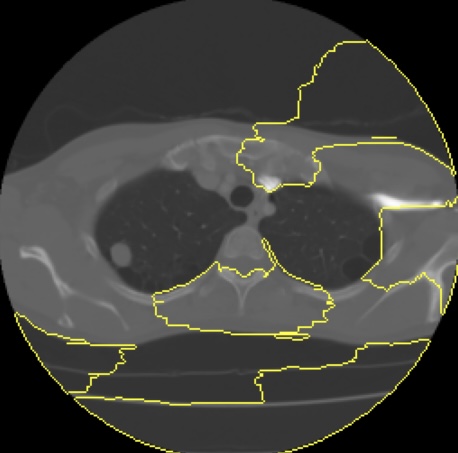}
        {\small (b) Incorrect Classification}\par\medskip
        \label{fig:Resnet50_False_Case}
    \end{minipage}
    \vspace{2mm}
    \caption{Correct and Incorrect Predictions by ResNet50.}
    \label{fig:ResNet50_Case}
\end{figure}

\begin{figure}[H]
    \centering
    \begin{minipage}{0.42\textwidth}
        \centering

        \includegraphics[width=\textwidth]{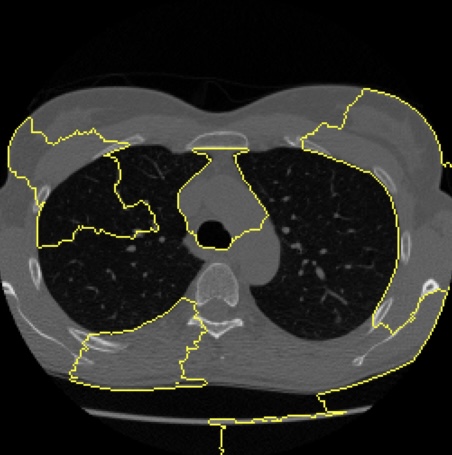}
        {\small (a) Correct Classification}\par\medskip
        \label{fig:ViT_Correct_Case}
    \end{minipage}
    \hspace{0.05\textwidth} 
    \begin{minipage}{0.39\textwidth}
        \centering
        
        \includegraphics[width=\textwidth]{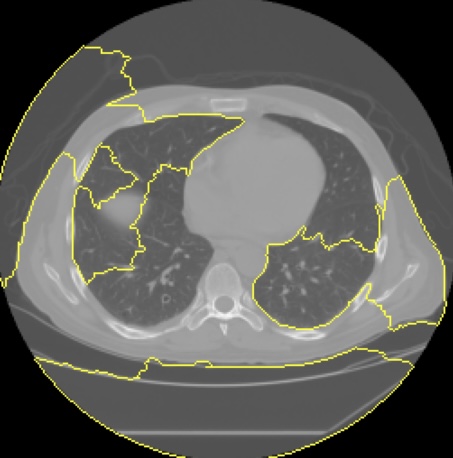}
        {\small (b) Incorrect Classification}\par\medskip
        \label{fig:ViT_False_Case}
    \end{minipage}
    \vspace{2mm}
    \caption{Correct and Incorrect Predictions by Vision Transformer.}
    \label{fig:ViT_Case}
\end{figure}

\subsection{Prediction Agreement vs LIME Explanation Agreement}
Pearson correlation was computed separately over the softmax prediction vectors and the LIME explanation vectors of each model pair to examine whether the models that are trained on similar conditions can agree on their outputs and also rely on similar image features.

\begin{table}[H] 
\centering
\caption{Prediction Probability Correlation Between Models}
\label{tab:prediction_correlation}
\setlength{\tabcolsep}{4pt}       
\renewcommand{\arraystretch}{1.2}  
\begin{tabular}{lccc}
\hline
\textbf{Models} & \textbf{Custom CNN} & \textbf{ResNet50} & \textbf{ViT} \\ \hline
Custom CNN & 1.00 & 1.00 & 0.99 \\ 
ResNet50   & 1.00 & 1.00 & 0.99 \\ 
ViT        & 0.99 & 0.99 & 1.00 \\ \hline
\end{tabular}
\end{table}

\begin{table}[H] 
\centering
\caption{LIME Explanation Correlation Between Models}
\label{tab:lime_correlation}
\setlength{\tabcolsep}{4pt}       
\renewcommand{\arraystretch}{1.2}  
\begin{tabular}{lccc}
\hline
\textbf{Models} & \textbf{Custom CNN} & \textbf{ResNet50} & \textbf{ViT} \\ \hline
Custom CNN & 1.00 & 0.25 & 0.22 \\ 
ResNet50   & 0.25 & 1.00 & 0.24 \\ 
ViT        & 0.22 & 0.24 & 1.00 \\ \hline
\end{tabular}
\end{table}

Table \ref{tab:prediction_correlation} shows the prediction correlations were near perfect across all three pairs with each getting $r$ value of 0.99. These values confirm that all three architectures assign highly consistent agreement scores to the same images across the test set, despite their structural differences.

Table \ref{tab:lime_correlation} shows the LIME correlations were comparatively lower across all pairs: CNN and ResNet50 ($r=0.2535$), CNN and ViT ($r=0.2247$), and ViT and ResNet50 ($r=0.2442$). All three values fall in a low positive range, reflecting weak agreement in the superpixels each model prioritizes when making decisions.

\begin{figure}[H]
    \centering
    \begin{minipage}{0.46\textwidth}
        \centering
        \includegraphics[width=\textwidth]{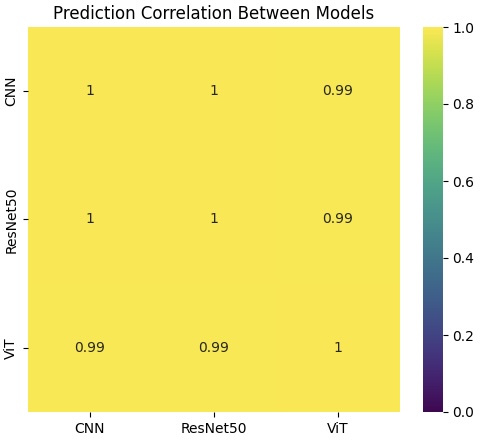}
        \par\medskip
        {\small (a) Prediction Correlation}
        \label{fig:pred_corr}
    \end{minipage}
    \hspace{0.05\textwidth} 
    \begin{minipage}{0.46\textwidth}
        \centering
        \includegraphics[width=\textwidth]{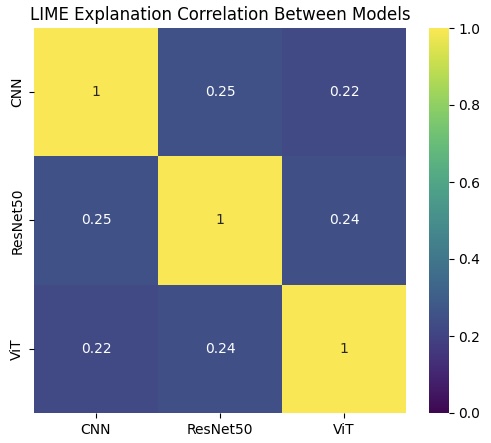}
        \par\medskip
        {\small (b) Explanation Correlation}
        \label{fig:explain_corr}
    \end{minipage}
    \vspace{2mm}
    \caption{Correlation heatmaps between Custom CNN, ResNet50, and Vision Transformer models.}
    \label{fig:correlation_heatmaps}
\end{figure}

\section{Discussion}

The central finding of this study is not the high classification accuracy by all three models, but rather the strong disagreement between their probability agreement and explanation agreement. Models coming to nearly similar conclusions at rate exceeding 0.99 shared less than 0.26 of their explanatory signal, meaning near identical diagnoses were reached through fundamentally different visual reasoning. The false alarm analysis deepens this concern. Incorrect predictions across all three architectures were consistently tied to attention outside the lung parenchyma, while correct predictions concentrated within it. This shows how model errors are locally structured rather than random, and that a model can be right for the wrong reasons in ways that accuracy metrics are structurally unable to expose. The outside lung attention bias observed in failure cases aligns with shortcut learning behavior documented widely in the medical imaging literature by Geirhos et al. \citep{geirhos2018imagenet, geirhos2020shortcut}. The present study extends these findings to a multi architecture CT setting and demonstrates that bias persists across convolutional and attention based architectures.

The prediction - explanation disagreement is consistent with the theoretical concerns raised by Adebayo et al. \citep{adebayo2018sanity}, who showed that model explanations can be visually reasonable without being statistically meaningful. However, while that work questioned the validity of individual explanations, the present findings go further by showing that even when explanations are taken at face value and compared across models, their inter model agreement is low despite near perfect output agreement. This is the novel contribution that existing literature has not addressed directly. Studies comparing CNN and ViT performance on medical imaging tasks, such as those by Gai et al. \citep{gai2024comparing}, have focused on accuracy differences and have not examined whether architecture divergence produces divergent feature attribution patterns. The low LIME correlations between models reported here fill that gap.

The practical implications of this study are significant. Prediction agreements across models, commonly treated as a signal of confidence in multi model voting systems, cannot be taken as evidence of reasoning consistency. For clinical deployment, attention maps should be evaluated against anatomical boundaries as a formal validation step rather than an optional supplement to accuracy reporting. A model cleared on performance benchmarks but failing this check should not be considered clinically relevant, regardless of its accuracy.

\section{Conclusion}
This study demonstrates that three models looked at the same CT scans, agreed on their answers nearly perfectly, and yet were paying attention to completely different parts of the image. That gap, between what models output and what they actually rely on, is what this work has tried to make visible. The fact that prediction correlations higher than 0.99 can coexist with explanation correlations less than 0.26 is not a statistical anomaly. It serves as a reminder that a consensus of models does not ensure safety. The real risk is in the clinical gap between the real world and the models, where the models fail despite passing all benchmark tests. The false alarm patterns that LIME identifies are precisely the kind of signals that are hidden by aggregate accuracy data, but they are arguably more crucial to patient safety than the data itself. This study is admittedly limited by its single site data and single explainability method, and both constraints are worth addressing in follow-up work through multi institution datasets and a broader set of attribution techniques. The bigger question this work leaves open is whether training models to look in the right places from the start, rather than auditing them afterward, could close the gap between prediction performance and genuine diagnostic reasoning.


\bibliographystyle{unsrtnat}
\bibliography{references}

@article{sung2021global,
  title={Global cancer statistics 2020: GLOBOCAN estimates of incidence and mortality worldwide for 36 cancers in 185 countries},
  author={Sung, Hyuna and Ferlay, Jacques and Siegel, Rebecca L and Laversanne, Mathieu and Soerjomataram, Isabelle and Jemal, Ahmedin and Bray, Freddie},
  journal={CA: a cancer journal for clinicians},
  volume={71},
  number={3},
  pages={209--249},
  year={2021},
  publisher={Wiley Online Library}
}

@article{hattori2017importance,
  title={Importance of ground glass opacity component in clinical stage IA radiologic invasive lung cancer},
  author={Hattori, Aritoshi and Matsunaga, Takeshi and Takamochi, Kazuya and Oh, Shiaki and Suzuki, Kenji},
  journal={The Annals of Thoracic Surgery},
  volume={104},
  number={1},
  pages={313--320},
  year={2017},
  publisher={Elsevier}
}

@article{fu2019distinct,
  title={Distinct prognostic factors in patients with stage I non--small cell lung cancer with radiologic part-solid or solid lesions},
  author={Fu, Fangqiu and Zhang, Yang and Wen, Zhexu and Zheng, Difan and Gao, Zhendong and Han, Han and Deng, Lin and Wang, Shengping and Liu, Quan and Li, Yuan and others},
  journal={Journal of Thoracic Oncology},
  volume={14},
  number={12},
  pages={2133--2142},
  year={2019},
  publisher={Elsevier}
}

@article{pandian2022detection,
  title={Detection and classification of lung cancer using CNN and Google net},
  author={Pandian, R and Vedanarayanan, V and Kumar, DNS Ravi and Rajakumar, R},
  journal={Measurement: Sensors},
  volume={24},
  pages={100588},
  year={2022},
  publisher={Elsevier}
}

@article{shariff2025optimizing,
  title={Optimizing non small cell lung cancer detection with convolutional neural networks and differential augmentation},
  author={Shariff, Vahiduddin and Paritala, Chiranjeevi and Ankala, Krishna Mohan},
  journal={Scientific Reports},
  volume={15},
  number={1},
  pages={15640},
  year={2025},
  publisher={Nature Publishing Group UK London}
}

@article{dosovitskiy2020image,
  title={An image is worth 16x16 words: Transformers for image recognition at scale},
  author={Dosovitskiy, Alexey and Beyer, Lucas and Kolesnikov, Alexander and Weissenborn, Dirk and Zhai, Xiaohua and Unterthiner, Thomas and Dehghani, Mostafa and Minderer, Matthias and Heigold, Georg and Gelly, Sylvain and others},
  journal={arXiv preprint arXiv:2010.11929},
  year={2020}
}

@article{faizi2025deep,
  title={Deep learning-based lung cancer classification of CT images},
  author={Faizi, Mohammad Khalid and Qiang, Yan and Wei, Yangyang and Qiao, Ying and Zhao, Juanjuan and Aftab, Rukhma and Urrehman, Zia},
  journal={BMC cancer},
  volume={25},
  number={1},
  pages={1056},
  year={2025},
  publisher={Springer}
}

@article{gai2024comparing,
  title={Comparing CNN-based and transformer-based models for identifying lung cancer: which is more effective?},
  author={Gai, Lulu and Xing, Mengmeng and Chen, Wei and Zhang, Yi and Qiao, Xu},
  journal={Multimedia Tools and Applications},
  volume={83},
  number={20},
  pages={59253--59269},
  year={2024},
  publisher={Springer}
}

@article{zech2018variable,
  title={Variable generalization performance of a deep learning model to detect pneumonia in chest radiographs: a cross-sectional study},
  author={Zech, John R and Badgeley, Marcus A and Liu, Manway and Costa, Anthony B and Titano, Joseph J and Oermann, Eric Karl},
  journal={PLoS medicine},
  volume={15},
  number={11},
  pages={e1002683},
  year={2018},
  publisher={Public Library of Science}
}

@article{rajasekar2023lung,
  title={Lung cancer disease prediction with CT scan and histopathological images feature analysis using deep learning techniques},
  author={Rajasekar, Vani and Vaishnnave, MP and Premkumar, S and Sarveshwaran, Velliangiri and Rangaraaj, V},
  journal={Results in Engineering},
  volume={18},
  pages={101111},
  year={2023},
  publisher={Elsevier}
}

@article{hossain2022transfer,
  title={Transfer learning with fine-tuned deep CNN ResNet50 model for classifying COVID-19 from chest X-ray images},
  author={Hossain, Md Belal and Iqbal, SM Hasan Sazzad and Islam, Md Monirul and Akhtar, Md Nasim and Sarker, Iqbal H},
  journal={Informatics in Medicine Unlocked},
  volume={30},
  pages={100916},
  year={2022},
  publisher={Elsevier}
}

@article{kumar2024vision,
  title={Vision transformer based effective model for early detection and classification of lung cancer},
  author={Kumar, Arvind and Mehta, Ravishankar and Reddy, B Ramachandra and Singh, Koushlendra Kumar},
  journal={SN Computer Science},
  volume={5},
  number={7},
  pages={839},
  year={2024},
  publisher={Springer}
}

@article{ukwuoma2025enhancing,
  title={Enhancing histopathological medical image classification for Early cancer diagnosis using deep learning and explainable AI--LIME \& SHAP},
  author={Ukwuoma, Chiagoziem C and Cai, Dongsheng and Eziefuna, Ebere O and Oluwasanmi, Ariyo and Abdi, Sabirin F and Muoka, Gladys W and Thomas, Dara and Sarpong, Kwabena},
  journal={Biomedical Signal Processing and Control},
  volume={100},
  pages={107014},
  year={2025},
  publisher={Elsevier}
}

@article{wani2024deepxplainer,
  title={DeepXplainer: An interpretable deep learning based approach for lung cancer detection using explainable artificial intelligence},
  author={Wani, Niyaz Ahmad and Kumar, Ravinder and Bedi, Jatin},
  journal={Computer Methods and Programs in Biomedicine},
  volume={243},
  pages={107879},
  year={2024},
  publisher={Elsevier}
}

@article{tomassini2023cloud,
  title={On-cloud decision-support system for non-small cell lung cancer histology characterization from thorax computed tomography scans},
  author={Tomassini, Selene and Falcionelli, Nicola and Bruschi, Giulia and Sbrollini, Agnese and Marini, Niccolo and Sernani, Paolo and Morettini, Micaela and M{\"u}ller, Henning and Dragoni, Aldo Franco and Burattini, Laura},
  journal={Computerized Medical Imaging and Graphics},
  volume={110},
  pages={102310},
  year={2023},
  publisher={Elsevier}
}

@inproceedings{geirhos2018imagenet,
  title={ImageNet-trained CNNs are biased towards texture; increasing shape bias improves accuracy and robustness},
  author={Geirhos, Robert and Rubisch, Patricia and Michaelis, Claudio and Bethge, Matthias and Wichmann, Felix A and Brendel, Wieland},
  booktitle={International conference on learning representations},
  year={2018}
}

@article{geirhos2020shortcut,
  title={Shortcut learning in deep neural networks},
  author={Geirhos, Robert and Jacobsen, J{\"o}rn-Henrik and Michaelis, Claudio and Zemel, Richard and Brendel, Wieland and Bethge, Matthias and Wichmann, Felix A},
  journal={Nature Machine Intelligence},
  volume={2},
  number={11},
  pages={665--673},
  year={2020},
  publisher={Nature Publishing Group UK London}
}

@article{adebayo2018sanity,
  title={Sanity checks for saliency maps},
  author={Adebayo, Julius and Gilmer, Justin and Muelly, Michael and Goodfellow, Ian and Hardt, Moritz and Kim, Been},
  journal={Advances in neural information processing systems},
  volume={31},
  year={2018}
}

@misc{panboonyuen2026seeingisntbelievinganalysis,
      title={Seeing Isn't Always Believing: Analysis of Grad-CAM Faithfulness and Localization Reliability in Lung Cancer CT Classification}, 
      author={Teerapong Panboonyuen},
      year={2026},
      eprint={2601.12826},
      archivePrefix={arXiv},
      primaryClass={cs.CV},
      url={\url{https://arxiv.org/abs/2601.12826}}, 
}

@article{dwivedi2023explainable,
  title={An explainable AI-driven biomarker discovery framework for Non-Small Cell Lung Cancer classification},
  author={Dwivedi, Kountay and Rajpal, Ankit and Rajpal, Sheetal and Agarwal, Manoj and Kumar, Virendra and Kumar, Naveen},
  journal={Computers in Biology and Medicine},
  volume={153},
  pages={106544},
  year={2023},
  publisher={Elsevier}
}

@article{bloice2017augmentor,
  title={Augmentor: an image augmentation library for machine learning},
  author={Bloice, Marcus D and Stocker, Christof and Holzinger, Andreas},
  journal={arXiv preprint arXiv:1708.04680},
  year={2017}
}

@article{raghu2021vision,
  title={Do vision transformers see like convolutional neural networks?},
  author={Raghu, Maithra and Unterthiner, Thomas and Kornblith, Simon and Zhang, Chiyuan and Dosovitskiy, Alexey},
  journal={Advances in neural information processing systems},
  volume={34},
  pages={12116--12128},
  year={2021}
}
\end{document}